\documentclass{article}

\PassOptionsToPackage{numbers, compress}{natbib}

    \usepackage[preprint]{neurips_2025}

\usepackage[utf8]{inputenc} 
\usepackage[T1]{fontenc}    
\usepackage{hyperref}       
\usepackage{url}            
\usepackage{booktabs}       
\usepackage{amsfonts}       
\usepackage{nicefrac}       
\usepackage{microtype}      
\usepackage[dvipsnames]{xcolor}         
\usepackage{amsmath}
\usepackage{mathtools}
\usepackage{svg}
\usepackage{graphicx}
\usepackage{grffile}
\graphicspath{ {./figures/illustrations/} {./figures/plots/scores/} {./figures/plots/skill_scores/}}
\usepackage{caption}
\usepackage{subcaption}
\usepackage[export]{adjustbox}
\usepackage{wrapfig}
\usepackage{amsthm}
\usepackage{algorithm}
\usepackage{algorithmic}
\usepackage{listings}
\usepackage{fancyvrb}
\theoremstyle{plain}

\theoremstyle{definition}

\theoremstyle{remark}

\definecolor{usercolor}{RGB}{0, 100, 150}
\definecolor{assistantcolor}{RGB}{150, 50, 0}
\definecolor{thinkcolor}{RGB}{100, 100, 100}
\definecolor{gtcolor}{RGB}{0, 150, 0}

\lstdefinestyle{llmstyle}{
    basicstyle=\footnotesize\ttfamily,
    breaklines=true,
    frame=single,
    backgroundcolor=\color{gray!5},
    showstringspaces=false,
    tabsize=2,
    captionpos=b
}

\title{OLLM: Options-\textit{based} Large Language Models}

\author{%
  Shashank Sharma \\
  Department of Computer Science\\
  University of Bath\\
  \texttt{ss3966@bath.ac.uk} \\
  \And
  Janina Hoffmann \\
  Department of Psychology \\
  University of Bath \\
  \texttt{jah253@bath.ac.uk} \\
  \AND
  Vinay Namboodiri \\
  Department of Computer Science \\
  University of Bath \\
  \texttt{vpn22@bath.ac.uk} \\
}

\begin{document}

\maketitle

\begin{abstract}

We introduce Options LLM (OLLM), a simple, general method that replaces the single next-token prediction of standard LLMs with a \textit{set of learned options} for the next token, indexed by a discrete latent variable.
Instead of relying on temperature or sampling heuristics to induce diversity, OLLM models variation explicitly: a small latent space parametrizes multiple plausible next-token options which can be selected or searched by a downstream policy.
Architecturally, OLLM is a lightweight "plug-in" that inserts two layers: an encoder and a decoder, before the output head, allowing almost any pretrained LLM to be converted with minimal additional parameters.
We apply OLLM to a 1.7B-parameter backbone (only $1.56\%$ of parameters trainable) trained on OpenMathReasoning and evaluated on OmniMath.
The SOTA LoRA-adapted baselines peak at $51\%$ final answer correctness, while OLLM's option set allows up to $\sim 70\%$ under optimal latent selection.
We then train a compact policy in the latent space that emits latents to control generation.
Operating in a low-dimensional option space makes reward optimization far more sample-efficient and substantially reduces common misalignments (e.g., language switching or degenerate reasoning), as the policy is constrained to options learned during SFT.
Crucially, this alignment arises from model structure rather than additional KL or handcrafted alignment losses.
Our results demonstrate that optionized next-token modeling enhances controllability, robustness, and efficiency in math reasoning, and highlight latent-space policy learning as a promising direction for reinforcement learning in LLMs.

\end{abstract}

\vspace{-.4cm}

\section{Introduction}

Traditional LLMs model a single next-token distribution and rely on sampling heuristics (temperature, top-k/p) to produce diversity.
In contrast, we propose Options LLM (OLLM), which learns a small latent option space that indexes multiple plausible next-token alternatives at each generation step.
By making variation explicit in a compact latent manifold, OLLM enables a downstream policy to act in a low-dimensional space for controlled generation—rather than over the full vocabulary—improving sample efficiency and constraining exploration to behaviors seen during supervised fine-tuning (SFT).
OLLM architecture is simple and involves adding two modules before the last layer (lm\_head): encoder and decoder. With this small change, any existing LLM can be converted into an OLLM.
Prior work has used compact latent spaces for attribute control and high-quality sampling \cite{gu2022controllable, ding2023maclasa,huang2025jam} and studied how RLHF can induce capability drift and reward-hacking \cite{lin2023mitigating}; OLLM differs by (i) exposing per-token options, (ii) integrating a light encoder/decoder plug-in into the pretrained backbone, and (iii) training a compact policy in the latent space to select options.
However, most of prior works require additional text labels, such as sentiment or topic, but OLLM does not require any labels.

We apply this setting to a distilled Qwen2.5 Deepseek R1 1.7B parameter model (only $1.56\%$ trainable parameters), training on the OpenMathReasoning dataset and evaluating on the Omnimath dataset.
We observe that the original module with SOTA LoRA adapters achieves a peak performance of $51\%$ (final answer correctness).
In contrast, our model can theoretically reach $70\%$ under optimal selection of the latent variable, indicating that the categorical space can learn the variations in text.
We then train a small policy module that functions in the latent space using SFT.
The policy module outputs the appropriate latent variable to control the generation.
Using the latent space for controlling generations causes the output to be bound to the variations learned during SFT.

\section{Methodology}

\subsection{Motivation}

\begin{wrapfigure}{r}{0.35\textwidth}
  \centering
  \includegraphics[width=\linewidth]{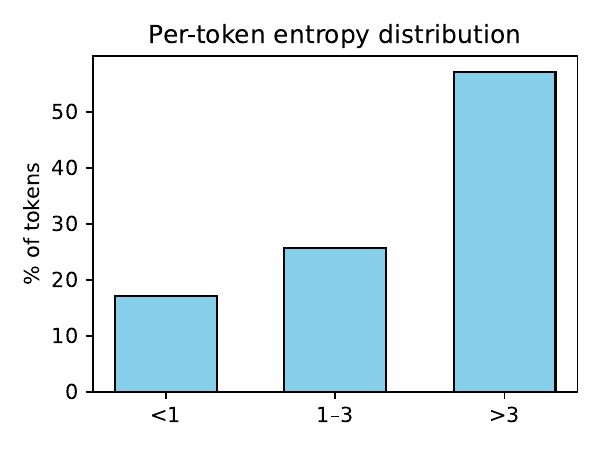}
  \caption{Percent of tokens grouped by their prediction entropy (deterministic, moderately non-deterministic, and highly ambiguous).}
  \label{fig:per_tok_entropy_grouped}
\end{wrapfigure}

Natural language generation often admits multiple plausible continuations at many token positions.
Empirically, we observed that $\sim 15\%$ of positions were deterministic (entropy $< 1$ nat), while $\sim 58\%$ admit high ambiguity (entropy $>3$ nats), confirming that next-token prediction often involves multiple plausible options (Fig. \ref{fig:per_tok_entropy_grouped}).
Moreover, the top-2 tokens account for less than $53.98\%$ of the probability mass, indicating genuine multimodality.
Conversely, a substantial fraction are effectively deterministic, e.g., token sequences with math expressions like "triangle $AB$" $\rightarrow$ "$\_C$", where the "$\_C$" is a near one-hot vector after the prefix.
Representative examples and full diagnostic tables are provided in Appendix \ref{sec:multimodal_probs}.
This mixture of ambiguous and deterministic sites motivates an explicit \textit{option} representation that can capture alternative next-token modes where they exist, while leaving deterministic positions unaffected.
Practically, modeling \textit{options} enables efficient downstream policy search (policy acts in a small action space).

\subsection{Architecture}

Let $x_{1:t}$ denote the observed prefix or context and $y_{t+1}$ the ground-truth next token.
A pretrained autoregressive model with parameters $\theta$ encodes the prefix into a hidden state $h_t \in \mathbb{R}^d$, which is normally mapped to logits $\ell_t = \mathrm{lm\_head}(h_t)$ over the vocabulary $V$.
To introduce options, we augment the model with two lightweight modules: an encoder $E_\psi$ and a decoder $D_\omega$.

During supervised training, the target token $y_{t+1}$ is first mapped through the model’s input embedding layer $e(\cdot)$ to obtain a compact representation $e(y_{t+1}) \in \mathbb{R}^d$.
The encoder then conditions jointly on this embedding and the contextual hidden state $h_t$ to produce a latent distribution (Eq.~\ref{eq:target_embed}),

\begin{equation}
    \label{eq:target_embed}
    p(z) = E_\psi\!\left(h_t, e(y_{t+1})\right) \in \mathcal{Z},
\end{equation}

where $\mathcal{Z}$ is a small discrete latent space of options.
This avoids scaling the encoder with the vocabulary size while still allowing it to condition on the identity of the target token.
And conditioning on both the token and its context prevents the encoder from collapsing into a static partition of the vocabulary, and instead allows it to learn context-sensitive latent assignments (e.g., the same target token may correspond to different options depending on the reasoning trajectory).

The decoder also takes $h_t$ as input, together with $z$, to generate a bias vector (Eq.~\ref{eq:decode_option}) that is added to the LLM hidden state before the vocabulary projection (Eq.~\ref{eq:final_hidden}).

\begin{gather}
    \label{eq:decode_option}
    \Delta h_t = D_\omega(z,h_t) \in \mathbb{R}^d \text{ where } z \sim p(z) \\
    \label{eq:final_hidden}
    \tilde h_t = h_t + \Delta h_t.
\end{gather}

\begin{figure}
    \centering
    \begin{subfigure}{0.46\textwidth}
        \centering
        \includegraphics[width=\textwidth]{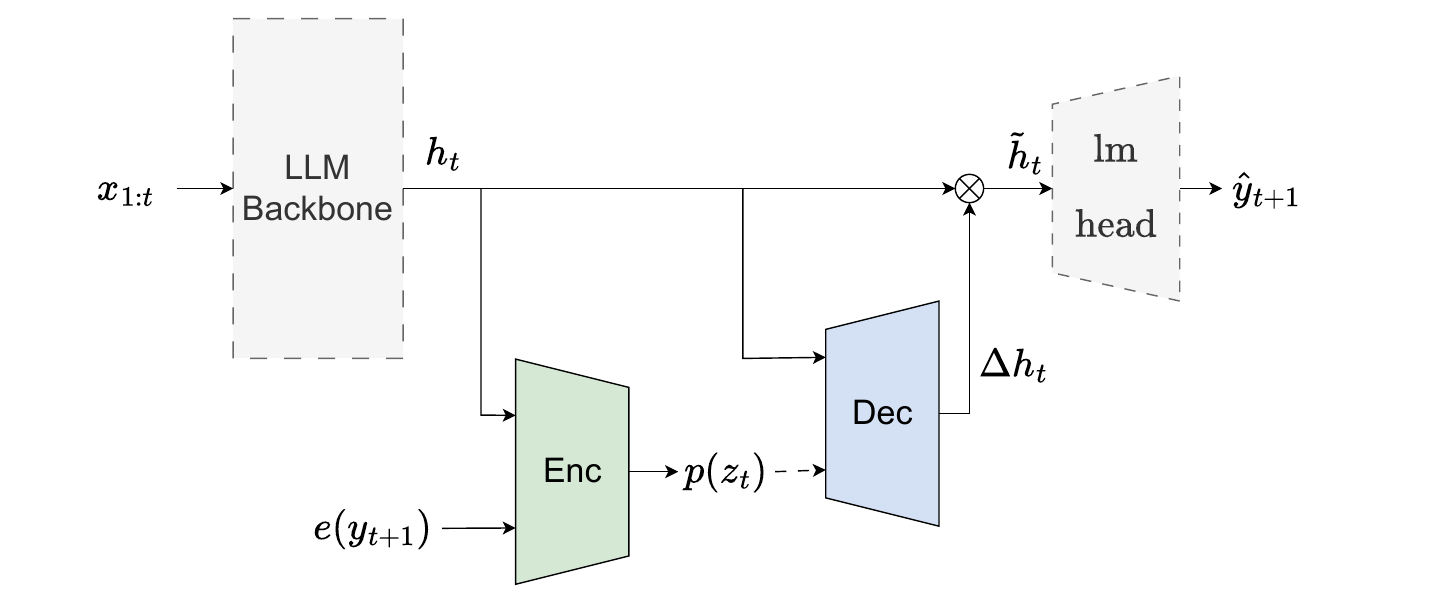}
        \caption{Training architecture}
        \label{fig:training_arch}
    \end{subfigure}
    \hspace{2em}
    \begin{subfigure}{0.46\textwidth}
        \centering
        \includegraphics[width=\textwidth]{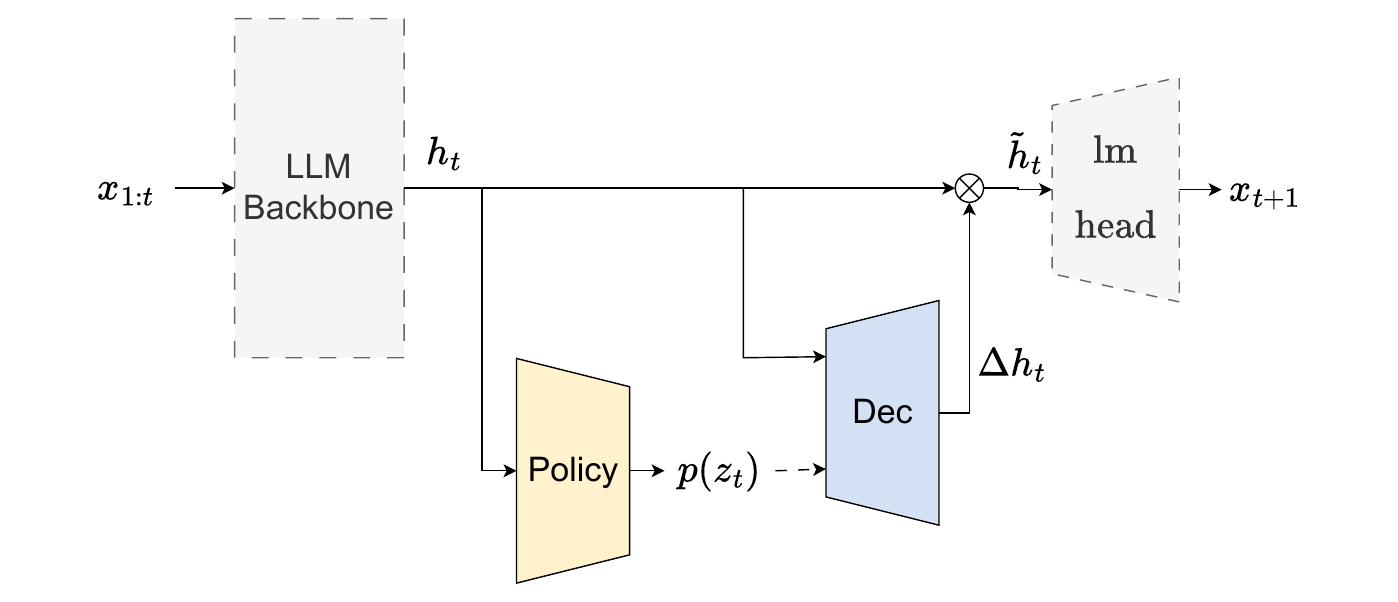}
        \caption{Inference architecture}
        \label{fig:inference_arch}
    \end{subfigure}
    \caption{Training and inference architectures for our method. The LLM backbone and the lm\_head remain fixed. (a) During training, the encoder predicts the appropriate latent by looking at the next token. (b) During training, a policy network predicts the appropriate latent. Policy is trained using SFT for behavioral cloning.}
    \label{fig:architecture}
\end{figure}

The logits are then computed as $\ell_t = \mathrm{lm\_head}(\tilde h_t)$, and training proceeds with the usual cross-entropy loss on $y_{t+1}$.
Figure \ref{fig:architecture} illustrates our architecture.

At inference, instead of the encoder, a lightweight policy $\pi_\phi(z \mid h_t)$ selects the appropriate latent using the hidden state $h_t$.
This design ensures that: (i) the backbone LLM is unchanged, (ii) only a small fraction of parameters (e.g., $1.56\%$ in our experiments) are trainable, and (iii) the action space for downstream optimization reduces to the latent set $\mathcal{Z}$ rather than the full vocabulary.

\begin{figure}
    \centering
    \includegraphics[width=\textwidth]{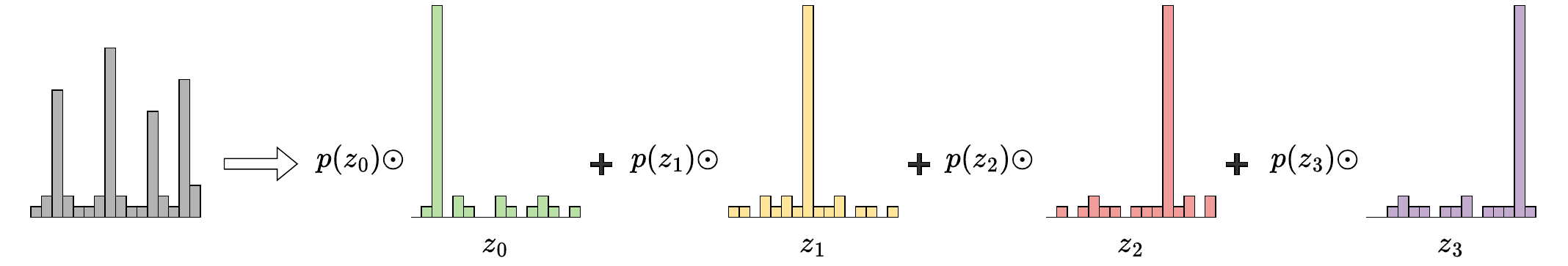}
    \caption{Illustration of the method impact. OLLM tries to decompose the token probabilities of the base model into a mixture of token probabilities using each categorical latent $z_i$. Thus, different target token options can be optimized against different latents.}
    \label{fig:prob_decompose}
\end{figure}

This formulation decomposes the next-token distribution into a mixture over latent options, $z \in \{1, \ldots, K\}$.
Each option allows predicting tokens a separate logit space, so competing continuations are disentangled: alternative tokens can be optimized under different latents rather than competing for probability mass within a single softmax (Fig. \ref{fig:prob_decompose}).

\subsection{Training Objectives}

\paragraph{Learning the option space.}
Given context $x_{1:t}$, the encoder $E_\psi$ outputs a probability distribution over the categorical latent $z \in \{1,\ldots,K\}$, ($K$-dim one-hot variable), and the decoder $D_\omega(z,h_t)$ biases the hidden state $h_t$ before projection.
The token prediction loss is the usual cross-entropy,

\begin{equation}
    \mathcal{L}_{\text{CE}}(\psi,\omega)
    = - \sum_t \log p(y_{t+1} \mid D_\omega(h_t, z_t)), 
    \quad z_t \sim E_\psi(h_t,e(y_{t+1})),
\end{equation}

ensuring that the correct token is predicted given the latent.
To avoid latent collapse, we regularize the encoder with a KL divergence to the uniform prior,

\begin{equation}
    \mathcal{L}_{\text{KL}}(\psi) = 
    \mathrm{KL}\!\left( E_\psi(\cdot \mid h_t,y_{t+1}) \; \| \; \mathcal{U}(1,K) \right).
\end{equation}

Rather than fixing the weight for KL loss, we adaptively rescale the KL term using an auto-adaptation module:
The scaling factor $\alpha$ is updated each step so that $\mathcal{L}_{\text{KL}}$ remains close to a target value (0.2 in our experiments).
This prevents both over-penalization in low-entropy regimes and under-penalization when the encoder distribution is peaked.

\paragraph{Policy learning.}
A lightweight policy $\pi_\phi$ is trained to imitate the encoder by predicting the latent $z$ from the hidden state $h_t$.
We use cross-entropy between the policy logits and the encoder’s latent assignment:

\begin{equation}
    \mathcal{L}_{\text{policy}}(\phi)
    = - \sum_t \log \pi_\phi(z_t \mid h_t), 
    \quad z_t = \arg\max E_\psi(h_t,y_{t+1}).
\end{equation}

The total loss is computed as: $\mathcal{L}(\psi,\omega,\phi) = \mathcal{L}_{\text{CE}}(\psi,\omega) + \mathcal{L}_{\text{KL}}(\psi) + \mathcal{L}_{\text{policy}}(\phi)$, and is optimized in one step.

\section{Results}

We use the OpenMathReasoning dataset \cite{moshkov2025aimo2} and evaluate using the Omnimath dataset \cite{gao2024omnimathuniversalolympiadlevel}, which contains math problems of comparable difficulty.
The latent space is set to $K=10$, allowing the model to represent up to ten token-level options at each step.

\begin{figure}
    \centering
    \begin{subfigure}{0.46\textwidth}
        \centering
        \includegraphics[width=\textwidth]{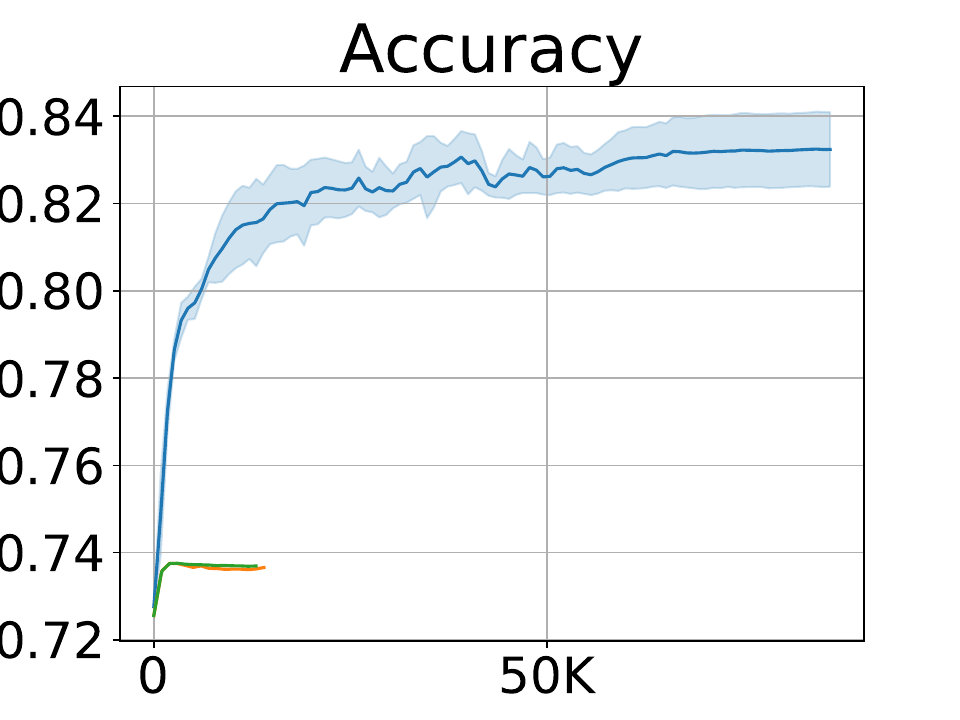}
        \caption{Per-token accuracy}
        \label{fig:eval_accuracy}
    \end{subfigure}
    \hspace{2em}
    \begin{subfigure}{0.46\textwidth}
        \centering
        \includegraphics[width=\textwidth]{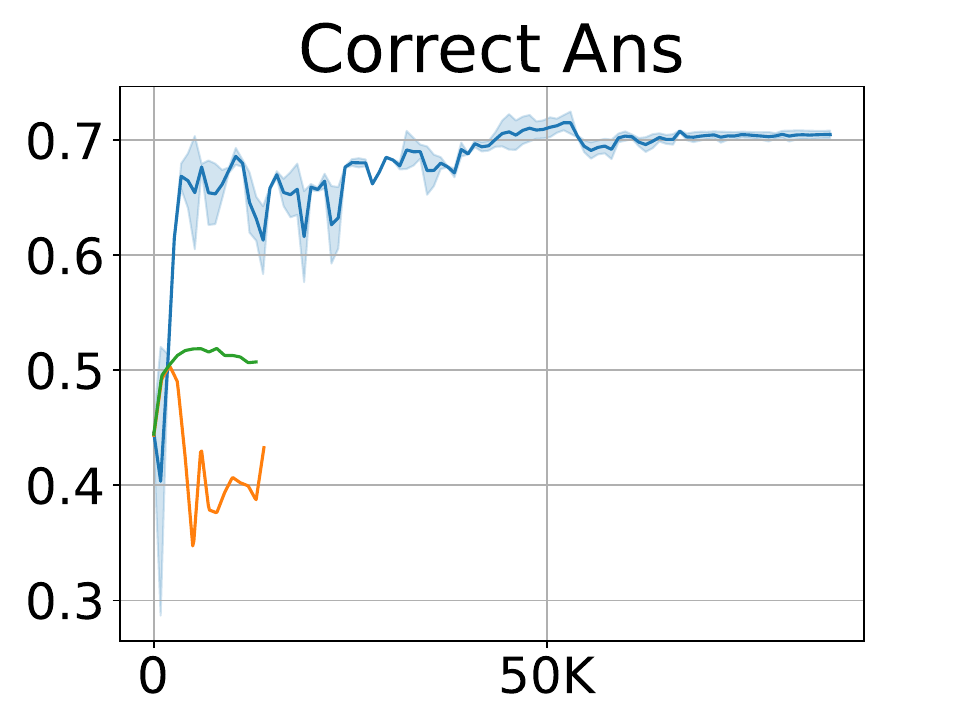}
        \caption{Final answer correctness}
        \label{fig:eval_correct}
    \end{subfigure}
    \begin{subfigure}{0.32\textwidth}
        \centering
        \includegraphics[width=\textwidth]{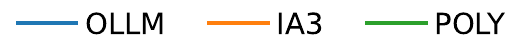}
    \end{subfigure}
    \caption{Evaluation curves comparing OLLM and state-of-the-art LORA modules. }
    \label{fig:eval_results}
\end{figure}

Figure~\ref{fig:eval_results} reports evaluation curves. Baseline LoRA-adapted models achieve per-token accuracy of $\sim 74\%$ and final-answer correctness of $\sim 51\%$, but plateau early due to overfitting. In contrast, OLLM continues improving throughout training. Under optimal latent selection, OLLM reaches $\sim 83\%$ per-token accuracy and $\sim 70\%$ final-answer correctness.

This gain arises because token alternatives do not compete within a single probability simplex: separating them into latent options allows the model to preserve multiple plausible continuations without overfitting to a dominant mode. Consequently, even simple policies that operate in the low-dimensional latent space can substantially improve final-answer accuracy.

We also provide qualitative examples of OLLM generations in Section~\ref{sec:ollm_samples}. These illustrate that OLLM learns genuinely diverse options when multiple continuations are plausible, but collapses to a consistent choice in deterministic cases (e.g., fixed symbol sequences). Notably, different options can also drive reasoning in distinct logical directions, underscoring the benefits of interpretability and controllability in the latent space.

\section{Discussion \& Future Work}

Our results show that introducing latent options into LLMs improves both token-level accuracy and final-answer correctness in mathematical reasoning. A key advantage is that the latent space provides a compact and well-grounded action space. Unlike traditional RL fine-tuning, which operates over the entire vocabulary and risks issues such as language switching or spurious reasoning, OLLM constrains exploration to options learned during supervised training. This removes the need for explicit KL penalties, reduces variance in policy learning, and ensures that exploration remains within valid continuations.

Looking ahead, we plan to train policies in this latent space using reinforcement learning, evaluating whether the theoretical efficiency gains translate into further improvements. The option structure also offers interpretability, as different latents can correspond to distinct reasoning trajectories, enabling controllable exploration of alternative solution paths. Extending OLLM to domains such as code generation, scientific QA, and multi-step planning represents a promising direction for future work.

\newpage
\bibliography{bibliography}


\newpage
\appendix

\section{Related Work}

A growing body of research has explored the use of latent spaces to enhance control, efficiency, and interpretability in large language models (LLMs). One common direction leverages latent probability transformations for controllable text generation. PriorControl \cite{gu2022controllable} and follow-up work on probability density estimation with normalizing flows \cite{dinh2016density} transform the complex, high-dimensional latent spaces of LLMs into structured Gaussian priors, enabling fine-grained single- and multi-attribute control. Similarly, compositional control via latent ODEs \cite{liu2022composable} shows that low-dimensional latent vectors permit efficient manipulation and composition of multiple operators (e.g., sentiment, tense, formality).

A second line of work investigates reinforcement learning and critic-driven methods in the latent or token distribution space. CriticControl \cite{kim2022critic} combines actor-critic RL with weighted decoding to manipulate token probabilities without modifying base model parameters. More recently, the EAGLE framework \cite{tennenholtz2024embedding} treats a pre-trained LLM as an environment and learns policies in the latent embedding space, demonstrating improved sample efficiency and alignment with domain-specific objectives.

Other approaches seek lightweight adaptation mechanisms for controllable generation. Prefix-tuning \cite{li2021prefix} and related parameter-efficient tuning methods optimize small continuous prompts or adapters in latent space, offering strong performance while preserving the base model. Recent multi-aspect control methods \cite{gu2022distributional} extend this by learning distributional intersections of attributes, reducing interference between controllers.

Finally, recent work has investigated interpretability and causal structure in latent representations. JAM (Just A Move) \cite{huang2025jam} frames text generation control as cause-effect reasoning in latent space, highlighting latent vectors as fundamental control units. These approaches complement ongoing efforts to better understand and guide the internal option-like structures of LLMs.

Our work is closest in spirit to these latent-space control frameworks but differs in its explicit formulation of \textit{options} as discrete latent variables aligned with next-token distributions. This design directly reduces competition among plausible tokens, enabling both improved reasoning accuracy and a natural foundation for downstream reinforcement learning in a compact action space.

\newpage
\section{Multimodality in Token distributions}
\label{sec:multimodal_probs}

\begin{figure}
    \centering
    \includegraphics[width=\textwidth]{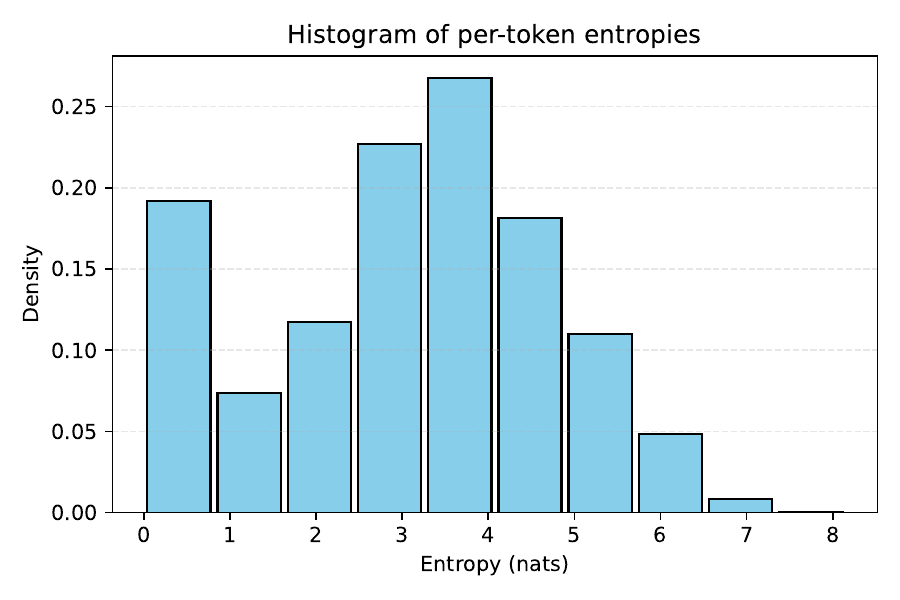}
    \caption{Histogram of the token entropies in predicted text for the OmniMath dataset. Two modes can be observed: one for near-deterministic tokens and another for tokens with multiple possible options.}
    \label{fig:per_tok_entropy}
\end{figure}

\begin{figure}
    \centering
    \includegraphics[width=\textwidth]{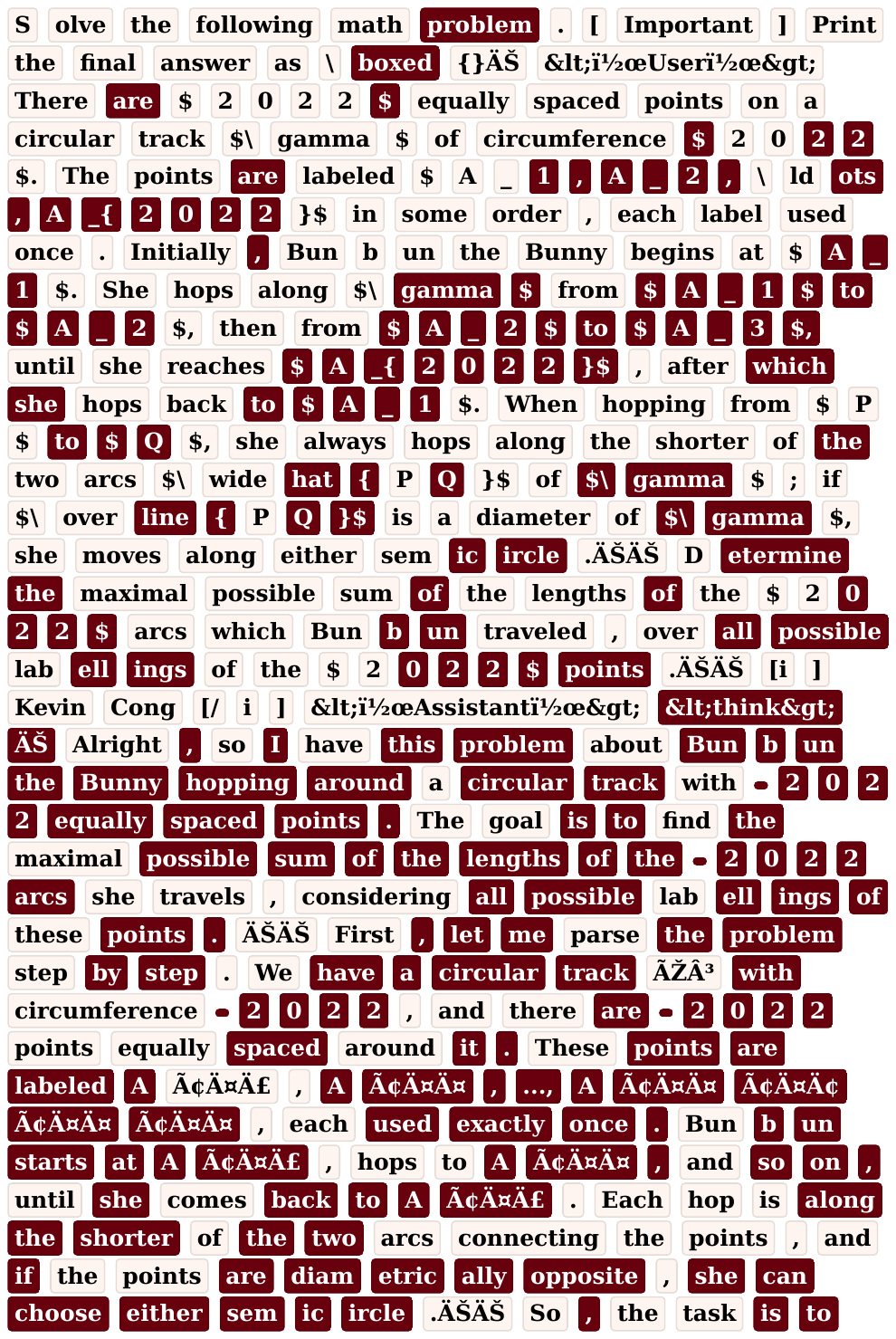}
    \caption{Example 1 showing the deterministic and ambiguous tokens. Tokens with probabilities $> .85$ are marked as deterministic (dark) and ambiguous (light) otherwise. Note how the start of phrases, sentences, or math expressions is ambiguous, but the endings are mostly deterministic.}
    \label{fig:example1}
\end{figure}

\begin{figure}
    \centering
    \includegraphics[width=\textwidth]{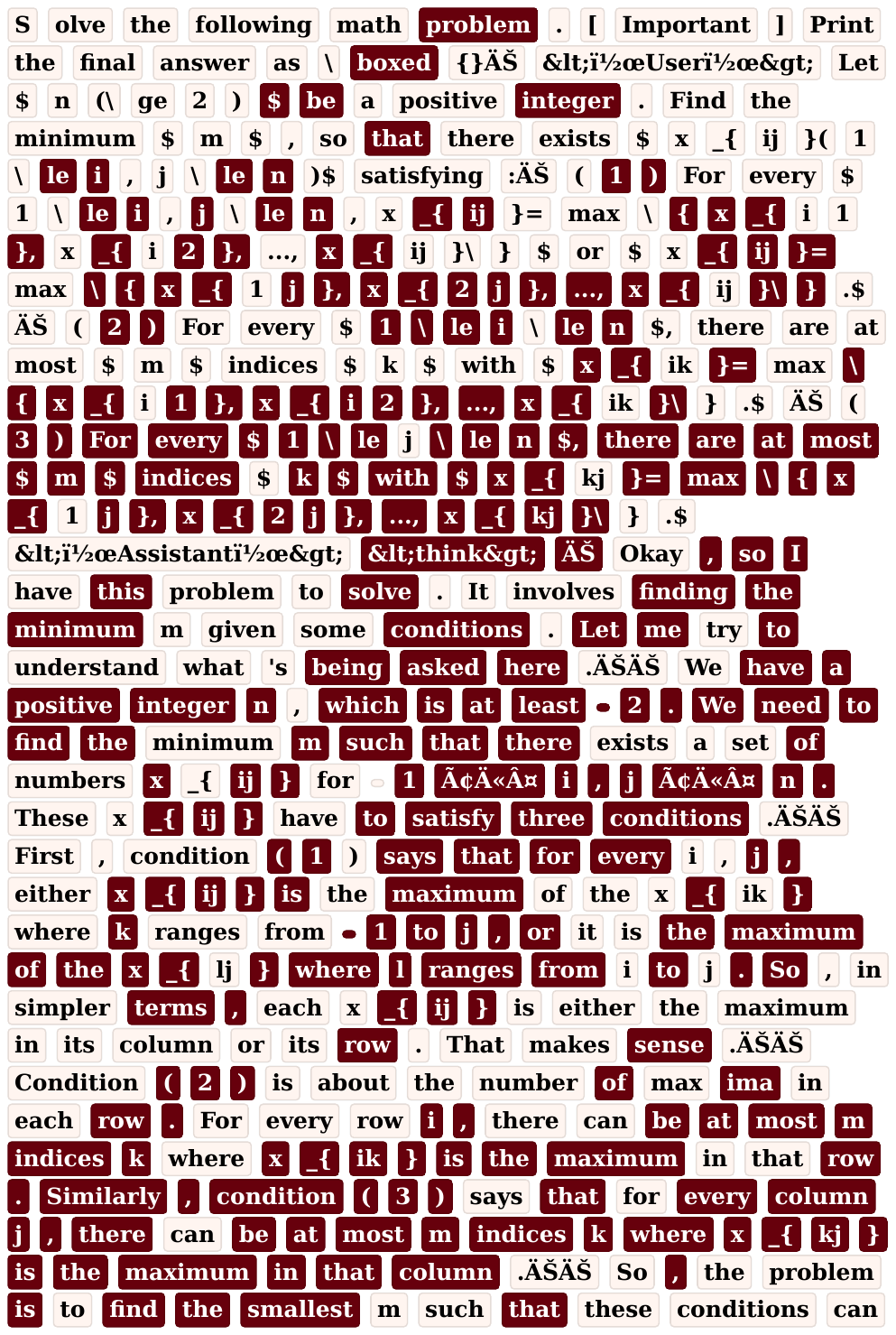}
    \caption{Example 2 showing the deterministic and ambiguous tokens.}
    \label{fig:example2}
\end{figure}

We examine the token probabilities of predictions for the Omnimath Dataset using the distilled Qwen2.5 Deepseek R1 1.5B parameter model.
We observed a significant proportion of both deterministic and non-deterministic tokens (Fig. \ref{fig:per_tok_entropy}).
Upon manually inspecting the individual examples, we found that non-deterministic tokens occur more frequently at the beginning of phrases, sentences, and expressions, but become deterministic as the generation progresses (Fig. \ref{fig:example1},\ref{fig:example2}).
This suggests that the LLM selects branches of logic initially through sampling and then follows through.

\newpage
\section{Samples from OLLM}
\label{sec:ollm_samples}

Examples of categorical options generated by OLLM for math problem solving. We vary the option latent and pick the token with the maximum probability. 'Opt prob' refers to the probability of the policy for the option.

\begin{figure}[htbp]
\begin{lstlisting}[style=llmstyle, caption={Here, the next word is certainly 'beans', so all options provide the same token.}]
<|begin_of_sentence|>Solve the following math problem. [Important] Print the final answer as \boxed{}

<|User|>Ken created a care package to send to his brother, who was away at boarding school.  Ken placed a box on a scale, and then he poured into the box enough jelly beans to bring the weight to 2 pounds.  Then, he added enough brownies to cause the weight to triple.  Next, he added another 2 pounds of jelly beans.  And finally, he added enough gummy worms to double the weight once again.  What was the final weight of the box of goodies, in pounds?

<|Assistant|><think>
Okay, let's see. So Ken is making a care package for his brother at boarding school. The box is on a scale, and Ken is adding various items to it. The problem says he first poured enough jelly beans to make the weight 2 pounds. Then he added brownies to triple the weight. After that, he added 2 pounds of jelly beans, and then he added gummy worms to double the weight again. We need to find the final weight of the box, right?

Hmm, so maybe breaking it down step by step would help. Let's try to figure out how much each item's weight adds and how the weight changes each time.

First, he adds jelly 

Opt  |  Opt prob  |  Token
  1  |  0.0639    |  beans
  2  |  0.1061    |  beans
  3  |  0.0996    |  beans
  4  |  0.1036    |  beans
  5  |  0.1394    |  beans
  6  |  0.0939    |  beans
  7  |  0.1069    |  beans
  8  |  0.1245    |  beans
  9  |  0.0544    |  beans
 10  |  0.1077    |  beans
\end{lstlisting}
\end{figure}

\begin{figure}[htbp]
\begin{lstlisting}[style=llmstyle, caption={Here, the next word has moderate ambiguity, so the token options are slightly varied.}]
<|begin_of_sentence|>Solve the following math problem. [Important] Print the final answer as \boxed{}
<|User|>Draw a $2004 \times 2004$ array of points. What is the largest integer $n$ for which it is possible to draw a convex $n$-gon whose vertices are chosen from the points in the array?<|Assistant|><think>
Alright, so the problem is: we have a 2004 x 2004 grid, which is like a 

Opt  |  Opt prob  |  Token
  1  |  0.0088    |  big
  2  |  0.5073    |  large
  3  |  0.0494    |  big
  4  |  0.0569    |  huge
  5  |  0.2800    |  big
  6  |  0.0158    |  big
  7  |  0.0100    |  big
  8  |  0.0526    |  chess
  9  |  0.0102    |  big
 10  |  0.0088    |  square
\end{lstlisting}
\end{figure}

\begin{figure}[htbp]
\begin{lstlisting}[style=llmstyle, caption={Here, the next word has high ambiguity, so the token options are even more varied.}]
<|begin_of_sentence|>Solve the following math problem. [Important] Print the final answer as \boxed{}
<|User|>Draw a $2004 \times 2004$ array of points. What is the largest integer $n$ for which it is possible to draw a convex $n$-gon whose vertices are chosen from the points in the array?<|Assistant|><think>
Alright, so the problem is: we have a 2004 x 2004 grid, which is like a large grid of points, right? 

Opt  |  Opt prob  |  Tokens
  1  |  0.0061    |  Let
  2  |  0.6528    |  And
  3  |  0.0780    |  Then
  4  |  0.0320    |  And
  5  |  0.0363    |  I
  6  |  0.1479    |  The
  7  |  0.0030    |  Each
  8  |  0.0278    |  So
  9  |  0.0053    |  Each
 10  |  0.0105    |   
\end{lstlisting}
\end{figure}

\begin{figure}[htbp]
\begin{lstlisting}[style=llmstyle, caption={Here, the next word can dictate which branch of logic the LLM can follow.}]
<|begin_of_sentence|>Solve the following math problem. [Important] Print the final answer as \boxed{}
<|User|>Given $30$ students such that each student has at most $5$ friends and for every $5$ students there is a pair of students that are not friends, determine the maximum $k$ such that for all such possible configurations, there exists $k$ students who are all not friends.<|Assistant|><think>
Hmm, let's try to figure out this problem. We have 30 students where each student has at most 5 friends. Also, every 5 students have at least one pair that aren't friends. We need to find the maximum k such that among any possible configuration, there exists k students who are all not friends. 

Alright, so first, let me parse the 

Opt  |  Opt prob  |  Tokens
  1  |  0.0033    |  problem
  2  |  0.0765    |  conditions
  3  |  0.7725    |  problem
  4  |  0.0158    |  problem
  5  |  0.0290    |  information
  6  |  0.0260    |  constraints
  7  |  0.0056    |  problem
  8  |  0.0046    |  problem
  9  |  0.0051    |  problem
 10  |  0.0615    |  question
\end{lstlisting}
\end{figure}

\end{document}